\documentclass[letterpaper, 10 pt, conference]{ieeeconf}
\IEEEoverridecommandlockouts                              % This command is only needed if 
\usepackage[colorinlistoftodos,obeyFinal]{todonotes}
\let\proof\relax

\usepackage{amsthm}
\usepackage{fainekos-macros}
\usepackage{amsmath} % assumes amsmath package installed
\usepackage{amssymb}  % assumes amsmath package installed
\usepackage{graphicx}
\usepackage{tikz}
\usetikzlibrary{arrows}
\usepackage{ifthen}
\usepackage{mathtools}
\usepackage{stmaryrd}
\usepackage{csquotes}
\usepackage{xargs}                      % Use more than one optional 
\usepackage{subcaption}
\usepackage{textcomp}
\usepackage{bm}
\graphicspath{{./figures/}}
\usepackage{algorithm}
\usepackage{algorithmic}
\usepackage{varwidth}
\usepackage{xspace}
\usepackage{multirow}

\def\eqref#1{equation~\ref{#1}}

\def\1{\bm{1}}

\def\topk{{\texttt{top}_k}}

\DeclareMathAlphabet{\mathsfit}{\encodingdefault}{\sfdefault}{m}{sl}
\SetMathAlphabet{\mathsfit}{bold}{\encodingdefault}{\sfdefault}{bx}{n}

\newcommand{\R}{\mathbb{R}}

\newcommand{\softmax}{\mathrm{softmax}}

\let\oldnl\nl% Store \nl in \oldnl
\newcommand{\nonl}{\renewcommand{\nl}{\let\nl\oldnl}}% Remove line number for one line

\newcommand\restr[2]{{% we make the whole thing an ordinary symbol
  \left.\kern-\nulldelimiterspace % automatically resize the bar with \right
  #1 % the function
  \littletaller % pretend it's a little taller at normal size
  \right|_{#2} % this is the delimiter
  }}

\usepackage{lipsum}

\usepackage{etoolbox}
\makeatletter
\patchcmd{\maketitle}{\@copyrightspace}{}{}{}
\makeatother

\usepackage{xcolor}

\usepackage{stmaryrd}
\usepackage{hyperref}
\usepackage{float,wrapfig}

\usepackage[acronym]{glossaries}
\glsdisablehyper
\newacronym{moe}{MoE}{Mixture of Experts}
\newacronym{sac}{SAC}{Soft Actor-Critic}
\newacronym[
longplural={Hybrid Contextual Markov Decision Processes}
]{hcmdp}{HcMDP}{Hybrid Contextual Markov Decision Process}
\newacronym[
longplural={Contextual Markov Decision Processes}
]{cmdp}{cMDP}{Contextual Markov Decision Process}
\newacronym{mpc}{MPC}{Model Predictive Control}
\newacronym{rl}{RL}{Reinforcement Learning}
\newacronym{sl}{SL}{Supervised Learning}
\newacronym{mhmoe}{MH-MoE}{Multi-Head Mixture-of-Experts}
\newacronym{ood}{OOD}{out-of-distribution}
\begin{document}

\title{\model: Reinforcement Learning with Mixture-of-Experts for Control of Hybrid Dynamical Systems with Uncertainty}

\author{Leroy D'Souza, Akash Karthikeyan, Yash Vardhan Pant, Sebastian Fischmeister% <-this % stops a space
\thanks{{Department of Electrical and Computer Engineering, University of Waterloo, Waterloo, Canada.}
{\tt\footnotesize \{l8dsouza, a9karthi, yash.pant, sfischme\}@uwaterloo.ca}.}%
\thanks{This work was supported in part by Magna International and the NSERC Discovery Grant.}
}

\maketitle
\begin{abstract}
Hybrid dynamical systems result from the interaction of continuous-variable dynamics with discrete events and encompass various systems such as legged robots, vehicles and aircrafts.  
Challenges arise when the system's modes are characterized by \emph{unobservable (latent) parameters} and the events that cause system dynamics to switch between different modes are \emph{also unobservable}. 
Model-based control approaches typically do not account for such uncertainty in the hybrid dynamics, while standard model-free RL methods fail to account for abrupt mode switches, leading to poor generalization.

To overcome this, we propose \model which models the actor of the \gls{sac} framework as a \gls{moe} with a learned router that adaptively selects among learned experts. To further improve robustness, we develop a curriculum-based training algorithm to prioritize data collection in challenging settings, allowing better generalization to unseen modes and switching locations.
Simulation studies in hybrid autonomous racing and legged locomotion tasks show that \model
outperforms baselines (up to 6x) in zero-shot generalization to unseen environments. Our curriculum strategy consistently improves performance across \emph{all} evaluated policies. Qualitative analysis shows that the interpretable \moe router activates different experts for distinct latent modes. \footnote{\url{https://ljdsouza.github.io/sacmoe/}}.
\end{abstract}
\glsresetall
\section{Introduction} \label{sec:introduction}
\gls{rl} algorithms are typically developed under the assumption of continuous, stationary system dynamics that are invariant to the environment that a system is operating in. 
However, many real-world systems exhibit \emph{hybrid} dynamics where the active dynamics model \emph{changes} due to spatially varying environmental conditions (e.g., a race car on an asphalt track with a wet patch at some location) and internal system state (e.g., different gaits for legged robots produce different joint angles that affect the system's contact dynamics~\cite{DHA}).
Such systems are called hybrid dynamical systems and designing \gls{rl} policies to control them remains a challenge.

\begin{figure}[t]
  \centering
  \includegraphics[width=0.3\textwidth,clip=true,trim={0 0 0 2cm}]{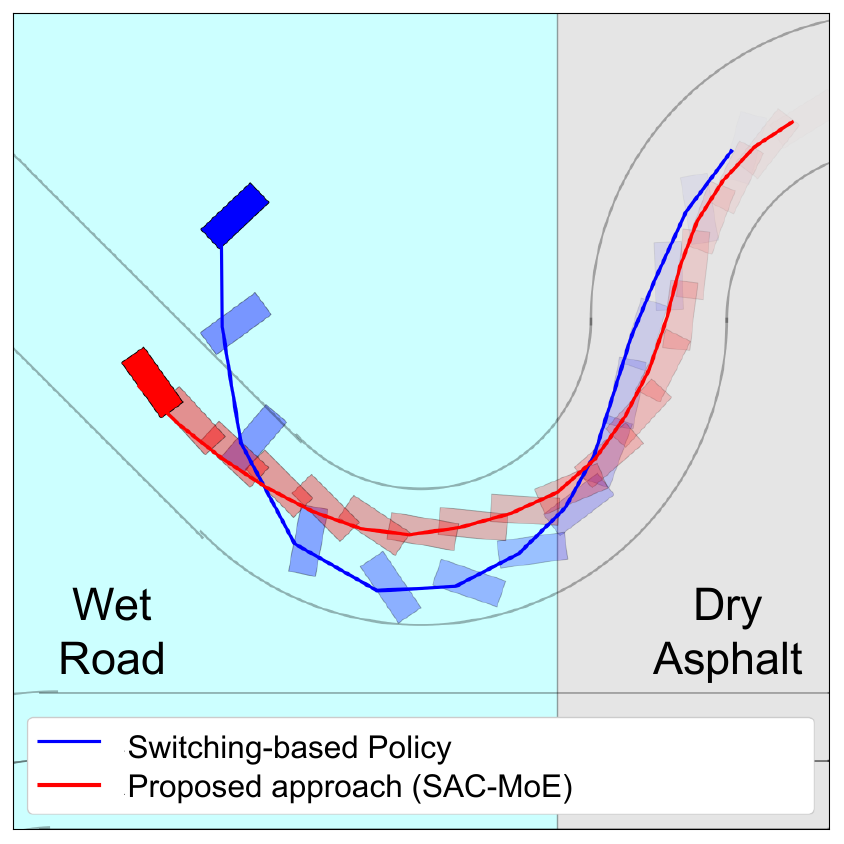}
  \caption{\small{Vehicle trajectories (right to left) in an environment with different road surfaces. The blue trajectory shows that switching between \emph{separate} policies for each mode results in failure. The red trajectory is from our proposed method, \model, which can successfully handle such transitions.}}
  \label{fig:motiv_fig}
  \vspace{-18pt}
\end{figure}

\begin{example}
\label{exmp:motiv}
Consider an autonomous vehicle navigating an environment with different terrains such as dry asphalt, wet surfaces, and snow. Each surface induces distinct dynamics due to differing friction, corresponding to different operating \emph{modes} of the hybrid system. As the vehicle transitions across terrains, its control strategy must be able to account for the changing dynamics during the transition. Figure~\ref{fig:motiv_fig} shows that switching between two policies learned \emph{separately} on asphalt and wet roads, may fail to handle transitions and skid off the track (blue trajectory). We propose a method that learns to account for such transitions between modes and better deals with hybrid system control (red trajectory). 
\end{example}

Example~\ref{exmp:motiv} indicates that we desire policies (controllers) for hybrid systems which account for \emph{discrete switching} (transitions) between different modes that each have their own dynamics. In this work, we consider the problem where i) modes are characterized, in part, by \emph{latent} (unobservable) parameters (e.g., friction coefficient) and, ii) mode switching locations are also \emph{a priori unknown}. Both these components can \emph{vary with the environment} and combine to form the (latent) hybrid system ``context'' (formalized in Section~\ref{sec:problem_prelim}). 

We desire policies that account for such latent context-dependent dynamics.
While the model-based control of hybrid systems~\cite{camacho_hmpc} has long been studied using optimization techniques, they usually consider piecewise affine systems~\cite{lazar_hmpc} and face tractability issues when working with general complex nonlinear dynamics~\cite{ACC23}. Moreover, they do not consider the challenges indicated above that result from a lack of context knowledge. This motivates the consideration of model-free \gls{rl} techniques to learn policies for hybrid systems that neither require such knowledge apriori nor suffer from tractability issues at inference time.

\noindent\textbf{Contributions of this work.} 
Motivated by the challenges described above, we now outline the key contributions of this work. We propose a) \model which parameterizes the actor in \gls{sac}~\cite{haarnoja2018soft} as a \gls{moe}~\cite{moe_1} to account for switching dynamics and robustly handle mode transitions using a \emph{learned router} that activates different (learned) experts, 
b) an adaptive curriculum learning scheme for hybrid systems that \emph{dynamically estimates} context hardness and prioritizes data collection in more challenging contexts to improve policy generalization across the space of contexts.

We evaluate our approach through extensive simulations on hybrid systems viz. autonomous racing across varied surfaces and legged locomotion. 
We show that the proposed \model approach outperforms the baselines by up to 6x in deployment settings with unseen contexts. We also show that the proposed curriculum approach consistently leads to better zero-shot generalization across \emph{both} the baselines and \model compared to other curricula demonstrating its general applicability.
Finally, we qualitatively demonstrate that the interpretable \moe router learns to select different experts for modes with very different dynamics while using similar experts for similar modes.

\section{Related Works}
\label{sec:related_works}

We briefly cover existing literature on the control of hybrid systems across both (model-free) \gls{rl} and model-based control domains.

\noindent\textbf{Reinforcement Learning for Hybrid System Control.}
Model-free RL has demonstrated impressive capabilities in tasks such as complex terrain traversal and legged locomotion. Hierarchical extensions, such as the option-critic architecture~\cite{bacon2017option}, enable end-to-end learning of policies over ``options,'' thereby allowing the discovery of specialized sub-policies for different terrains or gaits. However, such approaches are prone to degenerate behaviors, where a single sub-policy dominates the entire episode or terminates immediately upon switching and often require auxiliary signals
derived from domain knowledge~\cite{nachum2018data}.
Moreover, they usually assume access to ground-truth information (e.g., friction maps in Example~\ref{exmp:motiv}) to guide switching. In our setting, we assume this information is \emph{latent}, limiting the use of option-based methods in contrast to our proposed approach which employs a routing mechanism~\cite{moe_1,moe_2} to learn specialized sub-policies without access to latent parameters. 

The \moe architecture has been used in \gls{rl} for efficient training~\cite{ren2021probabilistic} and multi-task learning~\cite{cheng2023multi}, but has not been applied to hybrid system control. Our proposed approach adapts \moe to this setting, where mode parameters and switching locations are unobserved. Unlike prior work that can explicitly condition the router on task or context information, our formulation requires expert specialization to emerge implicitly, making it suitable for hybrid control problems of the type shown in Example~\ref{exmp:motiv}.

There exists work that applies \gls{rl} for hybrid system control by using discrete hybrid automata~\cite{DHA, NHA}. However, these do not consider transitions driven by environmental changes, a key challenge highlighted in Figure~\ref{fig:motiv_fig}.

\noindent\textbf{Control for \glspl{cmdp}.}
The hybrid systems we deal with can be framed as a \gls{cmdp} where the modes and mode switching describe the ``context'' (described in Section~\ref{sec:problem_prelim}).
It has been demonstrated that policies conditioned on \emph{observable} context can do better than ones without this information~\cite{contextualize_me} but may also perform suboptimally in unseen contexts~\cite{UPTrue}. In our setting, the context is \emph{unobservable} as indicated in Section~\ref{sec:introduction} preventing explicit policy conditioning on context information. AMAGO~\cite{amago} considers the problem of context-based meta-\gls{rl} using sequence models but does not consider switching between different dynamics modes of a hybrid system \emph{during} an episode. JCPL~\cite{jcplcontext_encoder} learns to encode context for policy conditioning. In our setting, the current and previously active modes in an episode do not necessarily tell us about what modes could be expected \emph{in the future}. This contrasts with JCPL's setting yielding a harder control task, particularly for zero-shot deployment.

\noindent\textbf{Model-based Control of Hybrid Systems.}
Model-based control-theoretic work on hybrid systems~\cite{hybrid_lit_review} has used Lyapunov theory for stability analysis, while most recent efforts employ optimization-based \gls{mpc}~\cite{camacho_hmpc}. Mode switching typically yields mixed-integer problems~\cite{borrelli2017predictive}, which are computationally demanding~\cite{ACC23, nlhybrid}. Data-driven \gls{mpc} methods extend this line by learning dynamics models for a fixed set of known modes, but require a priori mode-specific datasets~\cite{ecc2025_hgpmpc}. Such assumptions limit generalization to previously unseen modes. In contrast, we address the setting where mode parameters are latent, mode-specific data is not available a priori, and we desire effective generalization to unseen modes.

\section{Problem Statement}
\label{sec:problem_prelim}
We consider a hybrid system with state \( s_t \in \Sc \subset \mathbb{R}^n \), control input \( a_t \in \Ac \subset \mathbb{R}^u \), space of latent mode parameters $\Lc \subset \mathbb{R}^l$ with discrete-time dynamics of the form,
\begin{equation} \label{eq:hl_sys_dyn}
s_{t+1} \sim \sum_{m = 1}^{\envnummodes} \envswitch(m, s_t) f_m\left(s_t, a_t ; \mum\right) 
\end{equation}

\begin{figure}[t]
  \centering
  \includegraphics[width=0.75\columnwidth]{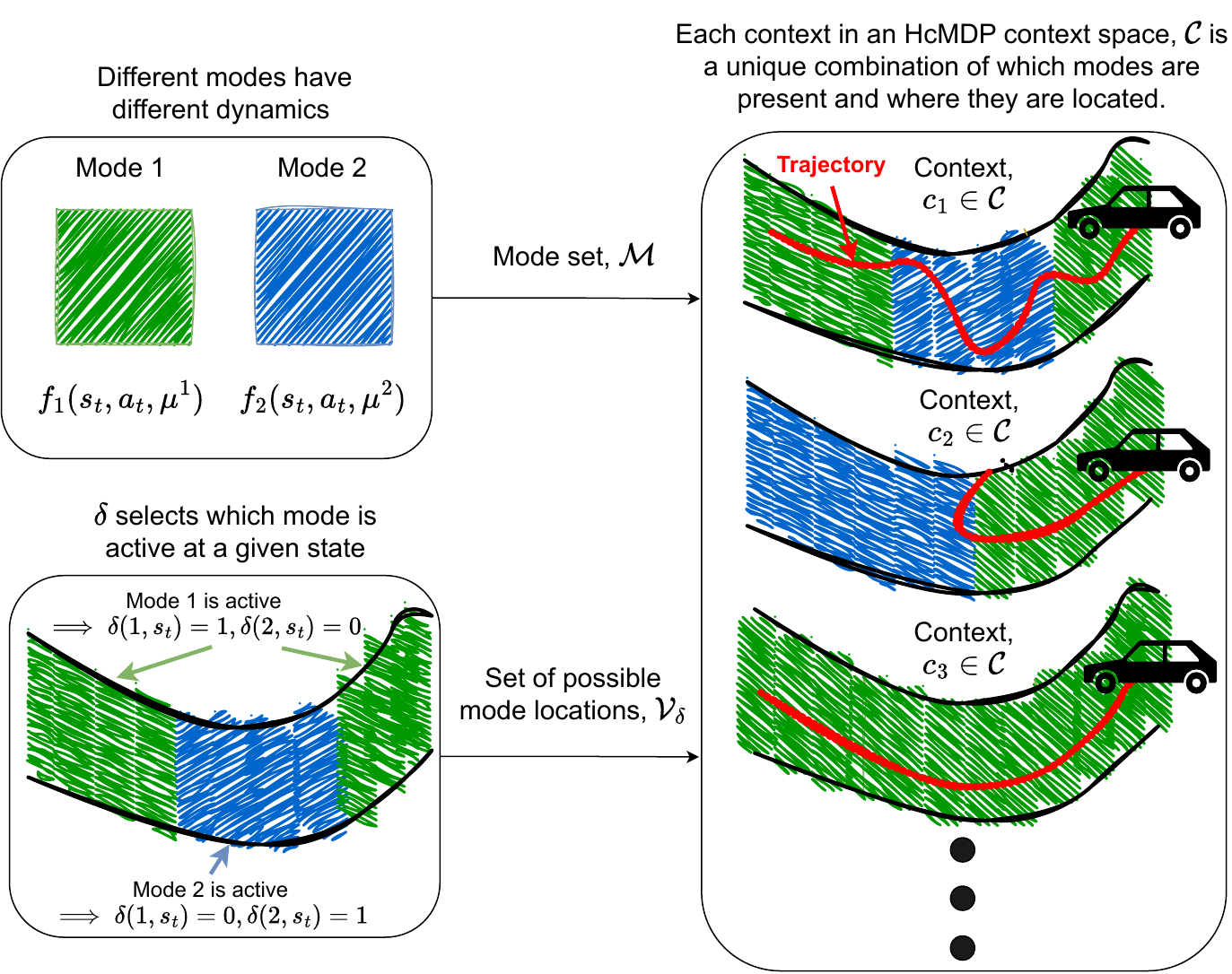}
  \caption{\small{A visualization of how mode parameters and mode location information are combined to generate various contexts that give rise to the \gls{hcmdp} context space, $\Ctrain$ in Definition~\ref{def:hcmdp}. Different contexts can lead to different trajectories for a given policy.}}
  \label{fig:hcdmp_var_viz}
  \vspace{-18pt}
\end{figure}

Here, $\envmodeset \subseteq \Lc$ denotes the set of latent mode vectors. \( f_m:\Sc \times \Ac \to \Delta(\Sc)\) denotes the stochastic dynamics associated with mode $m$, characterized by the latent parameter vector $\mum \in \envmodeset$, where $\Delta(\Sc)$ is a probability distribution over $\Sc$. The latent mode switching indicator, \( \envswitch(m, s_t): \envmodenumset \times \Sc \rightarrow \{0, 1\} \), determines which dynamics mode is active at a given state $s_t \in \Sc$. As indicated in Section~\ref{sec:introduction}, we define the context assignment, $c = (\envmodeset, \envswitch(\cdot, \cdot))$, which characterizes where different dynamics modes occur across the state space $\Sc$, including spatial variations across an environment (as in Example~\ref{exmp:motiv}). Thus, an environment is associated with a \emph{unique} context assignment (visualized in Fig.~\ref{fig:hcdmp_var_viz}), and we use the terms interchangeably henceforth.
Only one mode is active in a given state i.e., \( (\sum_{m=1}^{\envmodeset} \envswitch(m, s) = 1) \,\forall\, s \in \Sc\). We define $\vde$ as the space of valid switching indicators that satisfy this and use $\vdetrain \subset \vde$ to indicate the case where only a fraction of the possible mode switching locations are realized.

We now express these hybrid systems as a process over environments characterized by $\vdetrain, \Ltrain$ by adapting the definition of \glspl{cmdp}~\cite{hallak2015contextual}.
\glsunset{hcmdp}
\glsreset{hcmdp}
\begin{definition}
\label{def:hcmdp}
A \textbf{\gls{hcmdp}} is the tuple 
\((\Cc,\Lc,\vdetrain,\Sc,\Ac,R,\Tc)\), where \(\Sc\subset\mathbb R^n\), \(\Ac\subset\mathbb R^u\), \(\Lc\subset\mathbb R^l\) are the state, action and latent mode parameter spaces respectively. $\Cc = \{(\envmodeset, \envswitch) \mid \envmodeset \subseteq \Lc, \envswitch \in \vdetrain\}$ represents the space of (latent) contexts where each context $c \in \Ctrain$ corresponds to an environment characterized by the modes present in it (determined by $\envmodeset$) and the locations where these modes occur across $\Sc$ (determined by $\envswitch$). $\Tc(c)$ maps context $c \in \Cc$ to the environment-specific stochastic transition dynamics given by~\eqref{eq:hl_sys_dyn}. \(R:\Sc\times\Ac\to[r_{\min},r_{\max}]\) is the bounded reward function. 
\end{definition}

The combination of unknown mode dynamics with abrupt switching between modes that varies across environments makes learning a policy for controlling the hybrid system across different contexts a hard task. 
We consider learning a policy, $\pi_\theta(a|s): \Sc \rightarrow \Delta(\Ac)$ (where $\Delta(\Ac)$ is a parametric family of distributions over $\Ac$, that only obtains samples from a finite set of contexts ($\Cc$) during training and must generalize to previously unseen contexts ($\Ctest$) at test time. We formalize this in the problem statement below.

\begin{prob}\label{prob:hcmdp_pol_learning}
Given an \gls{hcmdp} \((\Ctrain, \Ltrain, \vdetrain, \Sc, \Ac, R, \Tc)\) defined by a finite set of latent modes, $\Ltrain$, and switching functions, $\vdetrain \subset \vde$, we aim to learn a control policy \( \pi_\theta(a \mid s) : \Sc \rightarrow \Delta(\Ac)\) that maximizes the expected return,
$J(\pi) = \mathbb{E}_{c \sim \Ctrain, \tau \sim \pi}\left[\sum_{t=0}^{\infty} \gamma^t r_t\right]$ where $\tau$ denotes trajectories sampled under $\pi$, $\gamma$ is the discount factor and $r_t$ is the reward at timestep $t$. At deployment, the \gls{hcmdp} \((\Ctest, \Ltest, \vdetest, \Sc, \Ac, R, \Tc)\) with $\Ctest \supset \Ctrain$ (since $\Ltest \supset \Ltrain, \vdetest \supset \vdetrain$) produces previously unseen contexts requiring the policy to generalize to maintain performance.
\end{prob}

\subsection{Why switching policies are insufficient for hybrid systems}
The hybrid structure in~\eqref{eq:hl_sys_dyn} naturally motivates a switching policy, $\pisw$, that selects among a set of component policies $\{\pi_m\}_{m=1}^{\envnummodes}$, each specific to a different mode i.e., $\pi_i$ is a policy \emph{specialized} for $f_i(s_t, a_t ; \mu^i)$ alone. Formally,
\begin{equation} \label{eq:switched_policy}
\pisw(a\mid s) = \sum_{m=1}^{\envnummodes} \envswitch(m, s_t) \pi_m(a\mid s)
\end{equation}
Thus, $\pisw$'s selection is based on the mode active at timestep $t$ as specified by $\envswitch(m, s_t)$. In Example~\ref{exmp:switching_policy} below, we show that even if an oracle were to provide $\pisw$ with access to $\envswitch(\cdot, \cdot)$ (which is otherwise unobservable), it still fails to generate desirable behavior across mode switches for a hybrid system. 

\begin{example}\label{exmp:switching_policy}
Consider the nominal dynamics of a kinematic bicycle model~\cite{commonroad_kinbic} with additional perturbation terms parametrized by a latent mode parameter $\mu \in \Lc = \{0, 6\}$. $\theta, v$ are the heading angle and velocity and $\text{a}_\text{long}, \psi$ are the input acceleration and steering angle.
\begin{equation}
\label{eq:kin_bicycle_model}
\begin{aligned}
    \dot{\theta} &= \dot{\theta}_\text{nominal} -\left(0.05\, e^{\psi} + 0.15\, \text{a}_\text{long} \tanh(\theta)\right) \mu \\
    \dot{v} &= \dot{v}_\text{nominal} -\left(0.1\, \text{a}_\text{long} \cos(\theta) + 0.3\, v\, \sin(\psi)\right) \mu
\end{aligned}
\end{equation}
\begin{figure}[t]
\centering
\includegraphics[width=0.4\textwidth, trim={0cm 0cm 0cm 0cm},clip]{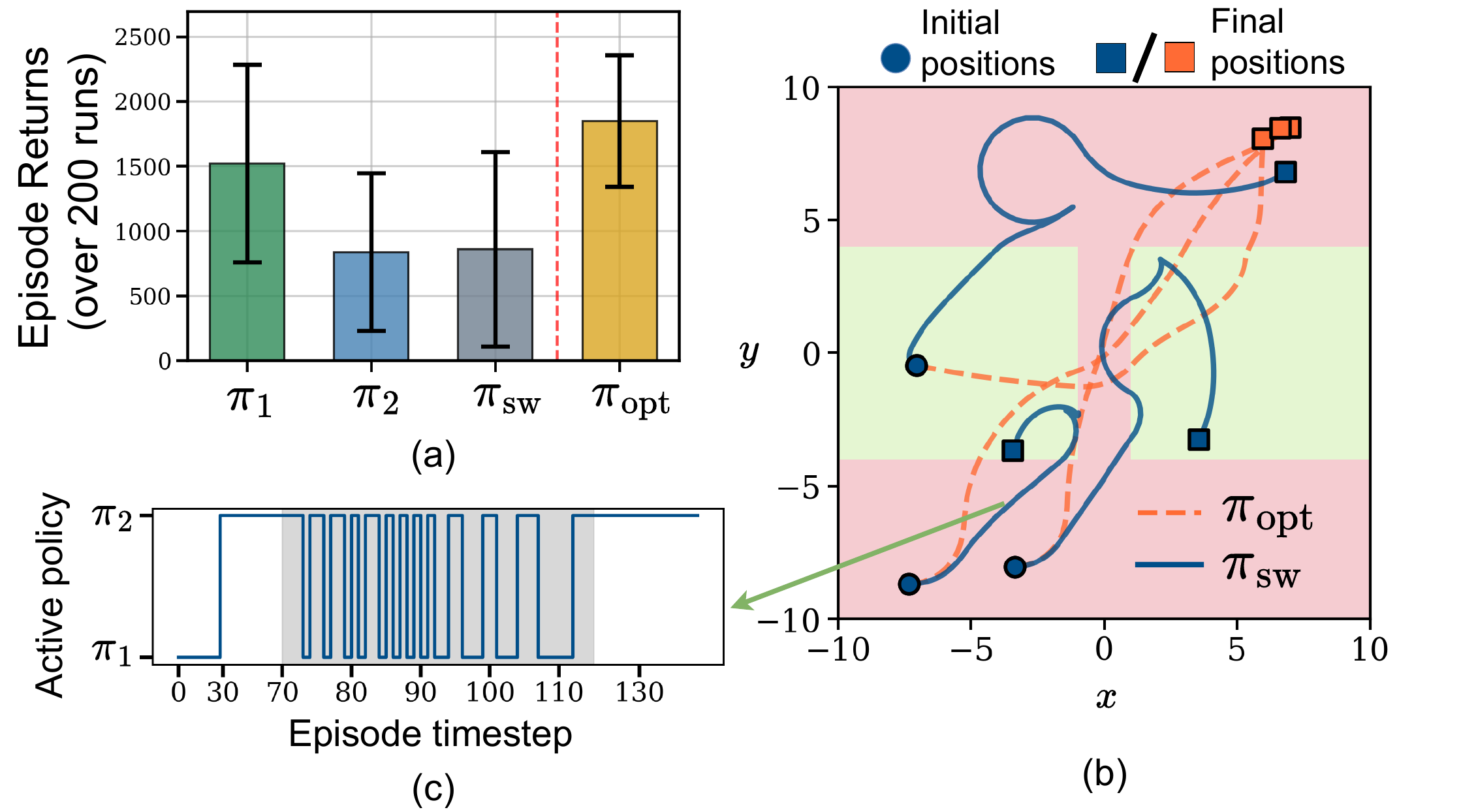}
\caption{\small{(a) Episode returns (goal-seeking reward) over 200 runs for policies in a test environment. (b) Visualization of the test environment (mode 1 in red and mode 2 in green) and selected trajectories for $\pisw$ and $\pi_\text{opt}$. (c) Visualization of $\pisw$ switching between component policies over a trajectory.}}
\vspace{-18pt}
\label{fig:hybrid_rl_switching_demo}
\end{figure}
Policies $\pi_1$ and $\pi_2$ are first trained separately in \emph{single-mode} environments, obtained by fixing 
$\mu=\mu^1=0$ and $\mu=\mu^2=6$ in~\eqref{eq:kin_bicycle_model}, respectively.   

Now, consider a test environment where both modes are present 
($\envmodeset = \{\mu^1, \mu^2\}$) and distributed spatially across the workspace. 
For reference, we also include $\pi_\text{opt}$, trained \emph{directly} in this test environment. 
Figure~\ref{fig:hybrid_rl_switching_demo}(a) shows that the switching policy $\pisw$, which selects between $\pi_1$ and $\pi_2$, achieves \emph{lower} return than even $\pi_1$ alone. 
As shown in Fig.~\ref{fig:hybrid_rl_switching_demo}(b), $\pisw$ produces suboptimal trajectories, and in the worst case (Fig.~\ref{fig:hybrid_rl_switching_demo}(c)) becomes trapped at mode boundaries, oscillating between conflicting policies, despite being provided with information from an oracle. 
\end{example}

Example~\ref{exmp:switching_policy} highlights that different component policies aim to generate different (possibly conflicting) conflicting behaviors \emph{over a trajectory} but are prevented from realizing their objectives due to the abrupt switching in $\pisw$. This leads to myopic and suboptimal behavior which Section~\ref{sec:results} shows to also manifest in other systems. This motivates the need for methods that learn policies \emph{not restricted to the form of the switching policy} in~\eqref{eq:switched_policy} to avoid these pitfalls.

\section{\model: Method}
\label{sec:method}
\begin{figure}[t]
\centering
\includegraphics[width=\columnwidth]{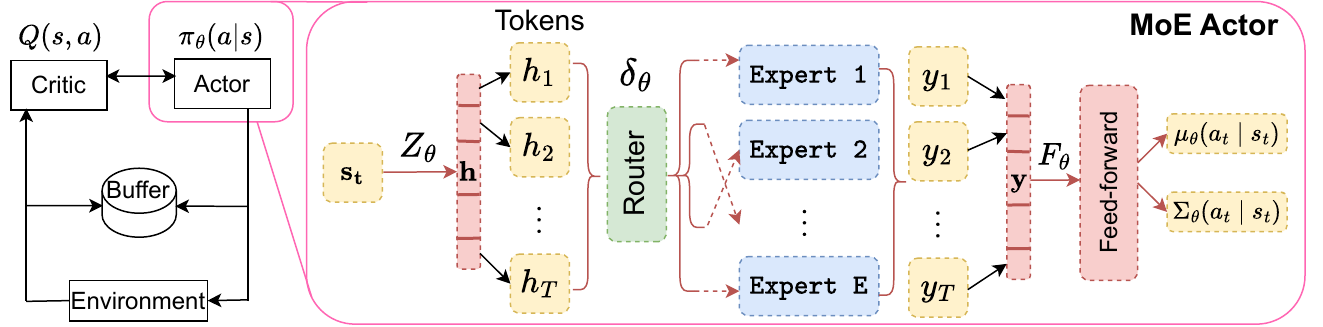}
\caption{\small{\textbf{\model{} Overview.} We adopt an actor-critic framework, where the actor is parameterized as a Mixture-of-Experts model. The encoder produces an embedding which is split into tokens. The router mechanism assigns tokens to different experts and merges their outputs to produce the action distribution.}}
\label{fig:method}
\vspace{-18pt}
\end{figure}

\glsreset{moe}
In this section, we propose \model{}, an actor–critic framework built on the \gls{sac} formulation~\cite{rl_1} with the actor being parameterized as a \gls{moe}, enabling it to handle mode switching in \gls{hcmdp} contexts and generalize to \emph{unseen} contexts at inference (Problem~\ref{prob:hcmdp_pol_learning}, Example~\ref{exmp:motiv}).  
This design addresses two central challenges: i) decomposing the policy into expert sub-policies that specialize without access to mode labels, and  ii) employing a sparse, learned router to adaptively compose experts based on the observed state, thereby improving generalization and control around mode transitions.  
We first detail the \moe actor in Section~\ref{subsec:moe_actor_rep} and compare it to the switching policy of Example~\ref{exmp:switching_policy}.  
We then show how the \moe actor integrates into SAC (Fig.~\ref{fig:method}) and present the overall training objective in Section~\ref{subsec:sacmoe_training}. Finally, Section~\ref{sec:curriculum_learning} introduces an adaptive curriculum algorithm that dynamically prioritizes learning on harder contexts, improving policy generalization while avoiding overfitting to easier contexts.
\subsection{Policy Representation} \label{subsec:moe_actor_rep}
Here, we define the components that make up our \moe actor as shown in Fig.~\ref{fig:method}.

\textbf{Tokenization.}  
Given a state $s$, an encoder $Z_\theta$ produces a latent representation $\mathbf{h} = Z_\theta(s)\,\in\,\R^{T\times d},$ which we split into $T$ tokens $\{h_i\}_{i=1}^T$, each of dimension $d$~\cite{wu2024multihead} allowing each token to capture distinct (complementary) aspects of the state for more holistic control.

\textbf{Experts.}  
We use a set of $E$ learnable experts, $\{\pithetae\}_{e=1}^E$, where each expert is a parameterized sub-policy that maps input tokens to outputs $\{\, \yie = \pithetae(h_i) \mid e \in \{1,\dots,E \,\}\} \,\forall\, i \in \{1,\dots,T\}$. 
In practice, different experts may learn to specialize on particular subsets of tokens which necessitates learning a \emph{weighted} combination of their outputs, as described below.

\textbf{Router.}  
The router $\delta_\theta$ assigns each token $h_i$ a probability distribution over the $E$ experts by producing logits that are transformed into routing probabilities $\alpha_i = \softmax(\delta_\theta(h_i)) \in \R^E$. We then select the $k$ experts with the highest routing probabilities for token $h_i$, denoted as $S_i = \topk(\alpha_i, k)$, and dispatch $h_i$ to those experts. The outputs of the selected experts $e \in S_i$ are weighted by their normalized routing score $\alphatildie = \tfrac{\alphaie}{\sum_{j \in S_i} \alpha_i^j}$. The aggregated output for $h_i$ is then, 
\[
y_i = \sum_{e \in S_i} \alphatildie \,\yie
\]  

\textbf{Action Distribution.} The token outputs $\{y_i\}_{i=1}^T$ are concatenated and passed through a feed-forward layer, $F_\theta$, that parameterizes the mean $\mu_\theta(a \mid s)$ and covariance $\Sigma_\theta(a \mid s)$ of the action distribution.  

\textbf{Comparing \moe and switching policies.}  
The resulting actor can be viewed as,  
\begin{equation} \label{eq:moe_policy}
\pi(a \mid s) = \sum_{e=1}^E \delta_\theta^e(Z_\theta(s))\, \pithetae(a \mid s),
\end{equation}
where $\delta_\theta^e(Z_\theta(s))$ denotes the learned routing weight for expert $e$.  
This form resembles the switching policy in~\eqref{eq:switched_policy}, but with two key differences. i) In $\pisw$, $\envswitch(m, s_t)$ are \emph{binary} indicators that activate a single policy at a time. In contrast, $\delta_\theta^e$ are real-valued weights, enabling the router to \emph{compose} multiple expert sub-policies. ii) In $\pisw$, the policy selector and mode-specific policies are \emph{decoupled} as two separate components. In contrast, our \moe formulation has both the router and the experts are trained \emph{jointly} in an end-to-end framework.

These distinctions highlight that \moe policy is not just a relaxation of the switching policy, but has additional expressivity to allow adaptive specialization and behavior composition across latent modes inspite of not needing to condition the router on mode information.

\glsreset{sac}
\subsection{Policy Learning} \label{subsec:sacmoe_training}
\noindent We train our policy using the off-policy \gls{sac} framework~\cite{rl_1} with the actor parameterized using \moe described in Section~\ref{subsec:moe_actor_rep}. The actor's objective maximizes expected return and policy entropy,

{\small
\begin{equation}
\label{eq:sac_loss}
L_{\mathcal{A}}
\;=\;
\mathbb{E}_{s_t\sim\mathcal D,\, a_t\sim \pi_\theta(\cdot|s_t)}
\Big[
-\big(\min_{i=1,2} Q_{\psi_i}(s_t,a_t)\big)
-\alpha\,\mathcal H\!\left(\pi_\theta(\cdot|s_t)\right)
\Big]
\end{equation}}

where \(\theta\) denotes parameters from the encoder, router, and experts, \(\mathcal{H}(\pi_\theta(\cdot|s_t)) = - \mathbb{E}_{a_t \sim \pi_\theta} [\log \pi_\theta(a_t|s_t)]\) is the Shannon entropy of the policy, \(\alpha\) is the entropy temperature, and \(\mathcal{D}\) is the replay buffer. The critic \(Q_\psi(s,a)\) is trained with temporal-difference targets using double Q-learning~\cite{hasselt2010double}.  

To better capture a policy that can handle hybrid systems across the expert sub-policies, we include an auxiliary regularization load loss term to promote balanced expert usage and encourage expert specialization within the \moe router.

\textbf{Load Loss.}  
To regularize expert assignments, we adopt the \emph{load loss}~\cite{moe_2}, which penalizes variance in the number of tokens routed to different experts.  
For a mini-batch $\mathcal{B}$, we estimate the probability of each expert $e$ being selected for a token $b \in \mathcal{B}$ under noisy routing, defined as  
$p_e(b) = \mathbb{P}\!\left((W b)_e + \epsilon_{\text{new}} \geq \text{threshold}_k(b)\right)$,  
where $\epsilon_{\text{new}} \sim \mathcal{N}(0, 1/E^2)$ and $\text{threshold}_k(b)$ is the $k$-th largest router score for token $b$.  
The expected token assignment count for each expert is then $\text{load}_e(\mathcal{B}) = \sum_{b \in \mathcal{B}} p_e(b)$, and the load-balancing loss is given by the squared coefficient of variation,  
$L_{\text{load}}(\mathcal{B}) = \left(\tfrac{\mathrm{std}(\{\text{load}_e(\mathcal{B})\}_{e=1}^E)}{\mathrm{mean}(\{\text{load}_e(\mathcal{B})\}_{e=1}^E)}\right)^2$.  
Minimizing this term encourages more uniform expert usage, preventing collapse to a few experts and promoting stable specialization.

\textbf{Training Objective.}  
The overall training procedure alternates between updating the actor and critic. The actor is trained with the combined loss  
$
L(\theta) = L_{\mathcal{A}} + \lambda_{\text{load}}\ L_{\text{load}},
$
where \(L_{\mathcal{A}}\) is the standard \gls{sac} actor loss (Eq.~\ref{eq:sac_loss}), \(L_{\text{load}}\) is the load-balancing regularizer,  and \(\lambda_{\text{load}}\) is a weighting hyperparameter. The critic is updated using the Bellman residual with double Q-learning~\cite{hasselt2010double}.

\subsection{Curriculum Learning for \glspl{hcmdp}} \label{sec:curriculum_learning}

In \glspl{cmdp}, contexts are typically sampled from a \emph{static uniform} distribution, i.e., $c^i \sim \text{Uniform}(\Ctrain)$ at the start of each episode $i$~\cite{lmdp}. 
While this can be applied to \glspl{hcmdp}, it ignores the fact that some contexts can be more challenging than others. With reference to Example~\ref{exmp:motiv}, i) controlling the vehicle during an asphalt-to-ice transition is harder during a corner bend than a straight road due to lateral dynamics~\cite{commonroad_kinbic} and ii) controlling the vehicle during an asphalt-to-ice switch on a turn is harder than an asphalt-to-damp road switch since the dynamics shift is more drastic in the former case. Thus, both the mode switch locations (affected by $\envswitch$) and the modes present (affected by $\envmodeset$) govern the hardness of a context $c \in \Ctrain$. We propose a curriculum learning strategy~\cite{curriculum_survey} based on this insight.

\begin{remark} \label{rem:context_index}
To simplify the discussion, we talk about sampling $c^i \sim \Ctraindist$. However, we do not assume access to the context itself. $c^i$ should be interpreted as an \emph{index} into the finite training context set $\Ctrain$.
\end{remark}  

However, \glspl{hcmdp} present unique challenges for curriculum learning since, i) the relative hardness of contexts is not known \emph{a priori} and depends on the \emph{evolving} policy, ii) contexts may \emph{share} some modes and switching locations, but this structure is \emph{unobservable} to the curriculum, so such similarities cannot inform hardness estimates over $\Ctrain$. 
These challenges contrast with typical curriculum learning settings, where i) difficulty can be ordered using domain knowledge~\cite{rapid_loco}, or ii) the curriculum has \emph{access to context parameters} and can manipulate them directly to probe similarities between contexts~\cite{active_domain_randomization}, in contrast to our setting in  Remark~\ref{rem:context_index} where only elements of $\Ctrain$ can be sampled, \emph{not the underlying parameters}.

To address this, we develop the adaptive curriculum strategy in Algorithm~\ref{alg:curriculum_learning} that maintains a \emph{dynamic} sampling distribution $\Ctraindistdynamic$ over $\Ctrain$.  
Each episode produces an outcome $U$ (e.g., return), recorded for the sampled context $c^i$ (Line~\ref{algline:outcome}).  
A curriculum metric $g$ updates $\Gc[c^i]$, the empirical difficulty estimate of context $c^i$, using $U$ (Line~\ref{algline:hardness_est}).
Finally, a distribution function $d_g$ sets $\Ctraindistdynamic$ based on $\Gc$ to prioritize sampling of harder contexts (Line~\ref{algline:dist_update}).  
In Section~\ref{subsec:racetrack_results}, we instantiate this framework with specific choices of $U$, $g$, and $d_g$ to show how an adaptive curriculum can improve performance without requiring prior knowledge of hardness or latent mode structure.

\begin{algorithm}[t]
    \caption{Curriculum learning for training in a \gls{hcmdp}}
    \label{alg:curriculum_learning}
    \begin{algorithmic}[1]
        \STATE \textbf{Input:} Training context set $\Ctrain$, curriculum metric $g$, distribution function $d_g$
        \STATE Initialize $\Gc \gets \mathbf{0}_{|\Ctrain|}$
        \STATE Initialize $\Delta(\Ctrain) \gets \text{Uniform}(\Ctrain)$
        \FOR{$i = 1$ \TO $N_E$}
            \STATE $c^i \sim \Delta(\Ctrain)$
            \STATE $U \gets \mathrm{RunEpisode}(c_i)$  \hfill{\color{blue}{\texttt{\small{$\triangleright$ Record episode}}}} \label{algline:outcome}
            \STATE $\Gc[c^i] \gets g(\Gc[c^i], U)$  \hfill{\color{blue}{\texttt{\small{$\triangleright$ Hardness estimate of $c^i$}}}} \label{algline:hardness_est}
            \STATE $\Delta(\Ctrain) \gets d_g(\Gc)$  \hfill{\color{blue}{\texttt{\small{$\triangleright$ Context sampling dist.}}}} \label{algline:dist_update}
        \ENDFOR
    \end{algorithmic}
\end{algorithm}

\section{Simulation Studies}
\label{sec:results}
\begin{table*}[htbp]
\centering
\footnotesize
\setlength{\tabcolsep}{5pt}
\setlength{\fboxsep}{1pt}
\resizebox{0.95\textwidth}{!}{
\begin{tabular}{llcc|ccc}
\toprule
\textbf{Model} & \textbf{Curriculum} & \multicolumn{2}{c}{\textbf{Single Surface}} & \multicolumn{3}{c}{\textbf{Multi-Surface}} \\
 &  & \textbf{\texttt{\{High\}}} & \textbf{\texttt{\{Low\}}} & \textbf{\texttt{\{Low, Medium\}}} & \textbf{\texttt{\{Low, High\}}} & \textbf{\texttt{\{Low, Medium, High\}}} \\
\midrule
\multirow{1}{*}{\textbf{SAC-Sw}} & \textit{} & 0.62 / \colorbox{blue!20}{\textbf{13.2}} / 0.89 & \colorbox{red!20}{\textbf{2.76}} / \colorbox{blue!20}{\textbf{14.6}} / \colorbox{green!20}{\textbf{0.19}} & 0.03 / \colorbox{blue!20}{\textbf{14.1}} / 1.00 & - / - / 1.00 & - / - / 1.00 \\
\addlinespace[0.2em]
\multirow{1}{*}{\textbf{SAC-SM}} & \textit{C} & \colorbox{red!20}{\textbf{3.23}} / 14.4 / 0.11 & 1.26 / 15.2 / 0.64 & 0.03 / 16.0 / 1.00 & - / - / 1.00 & 0.02 / 15.0 / 1.00 \\
\addlinespace[0.2em]
\midrule
\multirow{3}{*}{\textbf{SAC}} & \textit{A} & 2.26 / \textbf{13.3} / 0.50 & - / - / \textbf{1.00} & - / - / 1.00 & - / - / 1.00 & - / - / 1.00 \\
 & \textit{B} & 2.81 / 13.9 / 0.29 & - / - / \textbf{1.00} & 0.02 / 16.5 / 1.00 & - / - / 1.00 & 0.08 / \textbf{14.9} / 1.00 \\
 & \textit{C} & \textbf{3.05} / 15.9 / \textbf{0.02} & - / - / \textbf{1.00} & \textbf{0.95} / \textbf{15.9} / \textbf{0.82} & \textbf{0.22} / \colorbox{blue!20}{\textbf{15.5}} / \textbf{0.97} & \textbf{1.06} / 15.9 / \textbf{0.73} \\
\addlinespace[0.2em]
\multirow{3}{*}{\textbf{SAC-UPTrue}} & \textit{A} & 2.53 / \textbf{13.4} / 0.49 & - / - / 1.00 & 0.01 / 15.8 / 1.00 & - / - / 1.00 & 0.03 / 15.2 / 1.00 \\
 & \textit{B} & 1.53 / 13.6 / 0.75 & 0.02 / 16.4 / 1.00 & 0.04 / \textbf{15.4} / 1.00 & - / - / 1.00 & 0.02 / \colorbox{blue!20}{\textbf{14.3}} / 1.00 \\
 & \textit{C} & \textbf{3.04} / 15.2 / \textbf{0.08} & \textbf{0.18} / \textbf{15.9} / \textbf{0.97} & \textbf{1.43} / 16.2 / \textbf{0.61} & \textbf{0.22} / \textbf{15.7} / \textbf{0.95} & \textbf{1.35} / 15.7 / \textbf{0.66} \\
\addlinespace[0.2em]
\multirow{3}{*}{\textbf{SAC-MoE (ours)}} & \textit{A} & \textbf{3.15} / \textbf{13.6} / 0.17 & - / - / 1.00 & 0.04 / 15.4 / 1.00 & - / - / 1.00 & - / - / 1.00 \\
 & \textit{B} & 2.93 / 14.2 / 0.28 & 0.05 / 16.4 / 1.00 & 0.03 / \textbf{15.3} / 1.00 & - / - / 1.00 & 0.09 / \textbf{15.0} / 0.99 \\
 & \textit{C} & 2.98 / 16.5 / \colorbox{green!20}{\textbf{0.00}} & \textbf{0.69} / \textbf{15.7} / \textbf{0.81} & \colorbox{red!20}{\textbf{2.32}} / 16.4 / \colorbox{green!20}{\textbf{0.24}} & \colorbox{red!20}{\textbf{1.37}} / \textbf{16.1} / \colorbox{green!20}{\textbf{0.67}} & \colorbox{red!20}{\textbf{2.03}} / 16.3 / \colorbox{green!20}{\textbf{0.40}} \\
\bottomrule
\end{tabular}
}
\caption{\small{\textbf{Quantitative racing results (Track 1).} Performance comparison across curricula and policies for varied contexts.
Entries show Laps Completed / Avg. Lap Time / Crash Rate. For contexts from a given set of evaluated surfaces, best results across all curricula for a given model are \textbf{bolded} and best results across all models are highlighted (\colorbox{red!20}{red}: laps ($\uparrow$), \colorbox{blue!20}{blue}: lap time ($\downarrow$), \colorbox{green!20}{green}: crash rate ($\downarrow$)).}}
\label{table:multi_curriculum_perf}
\vspace{-15pt}
\end{table*}

In this section, we evaluate and compare \model against baseline algorithms (Section~\ref{sec:baselines}) on two challenging hybrid control domains, i) autonomous racing~\cite{highwayenv} and ii) legged locomotion using the Walker2d-v5 in MuJoCo~\cite{towers2024gymnasium}. Both tasks involve latent mode dynamics and spatially varying mode transitions. Our experiments aim to answer the following questions,

\noindent\textbf{P1. Does adaptive curriculum learning in \glspl{hcmdp} improve performance?}  
We test whether dynamically prioritizing harder contexts (Algorithm~\ref{alg:curriculum_learning}) leads to better policy performance compared to uniform sampling.  

\noindent\textbf{P2. Can \model generalize to novel contexts?}  
At inference, policies encounter unseen contexts $\Ctest \supset \Ctrain$ (Problem~\ref{prob:hcmdp_pol_learning}). We examine whether the \moe actor improves generalization relative to baselines.  

\noindent\textbf{P3. Does the \moe router capture latent mode structure?}  
We analyze router selections over trajectories to assess if expert activations correlate with different latent modes and interpret behavior captured by the learned sub-policies.  

\subsection{Baselines}
\label{sec:baselines}
\noindent\textbf{SAC}~\cite{haarnoja2018soft}: Standard off-policy actor-critic with maximum entropy regularization.  \\
\textbf{\uptrue}~\cite{UPTrue}: SAC augmented with \emph{oracle access} to the active mode at every timestep. Unlike prior work~\cite{UPTrue}, we allow training in the full context space $\Ctrain$ rather than restricting learning to contexts with a \emph{single} latent mode. \\ 

\textbf{\sacsw (\sacswshort)}: A switching policy $\pisw$ (Example~\ref{exmp:switching_policy}) composed of policies trained \emph{separately} on single modes. During deployment, it is given \emph{oracle access} to the current active mode and selects the policy trained on the closest mode for control at each timestep.

SAC and the proposed \model do not observe mode labels during training or inference. For fair comparison, all methods are implemented in the stable-baselines3 framework~\cite{SB3}. Experiments use Python 3.11 on a 12-core CPU with an RTX A6000 GPU. For \model we use 8 heads, 4 experts, and noisy top-$k$ routing with $k=2$ and a load loss weight ($\lambda_\text{load}$) of $0.01$.

\subsection{Case Study 1: Autonomous Racing} \label{subsec:racetrack_results}
\begin{figure}[t]
    \centering
    \includegraphics[width=0.85\linewidth]{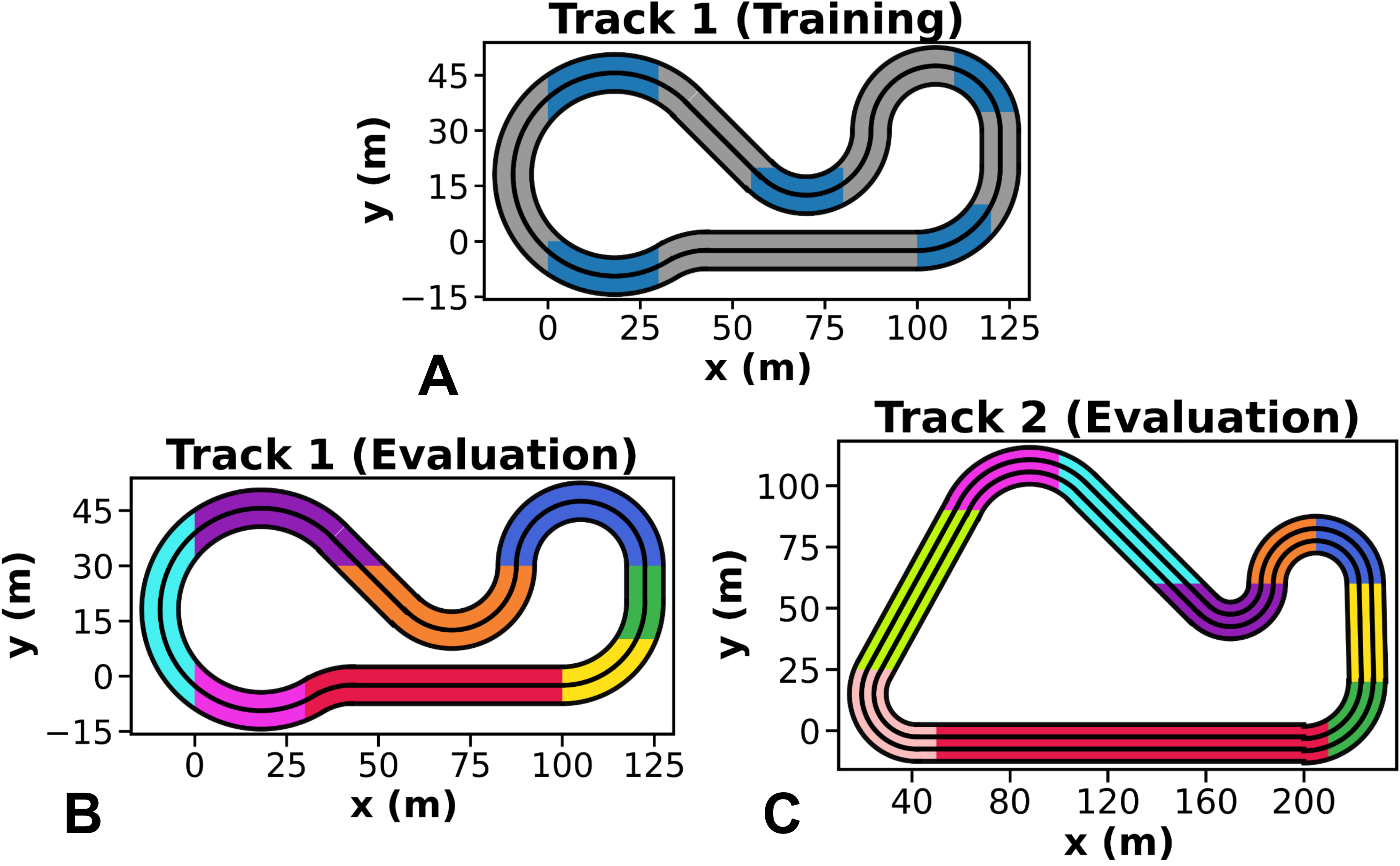}
    \caption{\small{\textbf{Autonomous racing setup.} \textbf{(A)} Training racetrack with $\Ctrain$ where each uniquely colored set of regions corresponds to a particular mode's (with value from $\Ltrain = \{1.0, 0.5, 0.3\}$) locations. \textbf{(B,C)} Evaluation tracks where colored regions represent surfaces with friction values sampled from predefined ranges, yielding diverse test contexts $\Ctest$. Track 2 is out-of-distribution.}}
    \label{fig:racing_track_viz}
    \vspace{-20pt}
\end{figure}

We first evaluate on an autonomous racing task where policies are incentivized to complete laps as quickly as possible without crashing. Hybrid dynamics arise from latent friction parameters, with mode switching induced by spatial variations in friction across the track (Fig.~\ref{fig:racing_track_viz}). Policies must therefore learn to handle abrupt friction changes, especially at high speeds and during cornering. We now make the training and testing \glspl{hcmdp} from Problem~\ref{prob:hcmdp_pol_learning} explicit.

\noindent\textbf{Training setup.}  
Training is performed on Track~1 with context space given by $\Ltrain = \{1.0, 0.5, 0.3\}$ (friction assignments) and $\delta \in \vdetrain$ as seen in Fig.~\ref{fig:racing_track_viz} \textbf{A}. The \sacswshort baseline consists of three component policies, one per friction value. For this section, we also define \textbf{SAC-SM}, trained like SAC but restricted to single-mode contexts i.e., $c = (\envmodeset, \delta) ~\text{s.t.}~ |\envmodeset|=1$. All policies are trained for 750K steps.  

\noindent\textbf{Evaluation setup.}  
At test time, we evaluate generalization to unseen contexts $\Ctest \supset \Ctrain$ across both Track~1 and Track~2 (out-of-distribution track layout) each partitioned into regions that yield $\delta \in \vdetest$ (exemplified in Fig.~\ref{fig:racing_track_viz} \textbf{B}, \textbf{C} respectively). Each region is randomly assigned a surface type that can have friction over a range of values viz., low $[0.25\!-\!0.35]$, medium $[0.45\!-\!0.6]$, or high $[0.8\!-\!1.0]$ friction. The union of these ranges yields the space $\Ltest$ of test latent mode parameters. This yields a much larger set of contexts $\Ctest$ than in training. Episodes are over 250 steps on Track~1 and 400 steps on Track~2.  

\noindent\textbf{Metrics.}  
We report average lap time, average number of laps completed without a crash, and crash rate, over 200 evaluation episodes per policy.  

\noindent\textbf{Curricula to Compare.}  
To address \textbf{P1}, we evaluate three curricula, with two derived from intuitive choices for the inputs to Algorithm~\ref{alg:curriculum_learning} to yield $\Ctraindistdynamic$,

\noindent\textit{Curriculum A (Uniform).} Static uniform sampling over $\Ctrain$, as in prior \gls{cmdp} work~\cite{lmdp}.  

\noindent\textit{Curriculum B (Step balancing).} $U=$ episode length. Contexts with fewer accumulated steps ($\Gc[c^i]$) are prioritized (by $d_g$), encouraging balanced context sampling.  

\noindent\textit{Curriculum C (Performance-aware).} $U=$ episode return. Context with lowest moving average return ($\Gc[c^i]$) is selected with probability $p=0.95$, and others are sampled randomly to avoid forgetting.  

\textbf{\textit{Result Discussion: }} Our results on Track~1 and Track~2 are summarized in Table~\ref{table:multi_curriculum_perf} and Table~\ref{table:newoodtrack_perf}, respectively. We structure the discussion in three parts, i) comparing different curricula (\textbf{P1}) ii) contrasting training in \glspl{hcmdp} with mode switches against \emph{single-mode} contexts and, iii) evaluating zero-shot generalization to unseen contexts (\textbf{P2}).

\noindent\textbf{Effect of Curricula (P1).}  
Curriculum~C consistently outperforms A and B across contexts (Table~\ref{table:multi_curriculum_perf}). A and B overfit to high-friction cases, failing to generalize across the remaining contexts. In contrast, C balances learning on harder (low return) contexts with random sampling to avoid forgetting. We therefore adopt Curriculum~C for all subsequent experiments.  

\noindent\textbf{Training in hybrid vs. single-mode contexts.}  
SAC-SM, restricted to training in single-mode contexts, specializes in single-surface evaluations but fails to adapt contexts with multiple surfaces that can have switches between drastically different modes. \sacsw, which selects among single-mode policies, exhibits similar limitations and even has component policies that overfit (e.g., the $\mu=1.0$ policy fails to generalize to the $[0.8,1.0]$ friction range). This confirms that training in single-mode contexts \emph{alone} is insufficient for control over general \gls{hcmdp} contexts.  

\noindent\textbf{Generalization performance (P2).}  
\model consistently outperforms the baselines in zero-shot generalization to novel contexts. On Track 1, it completes more laps with fewer crashes across most contexts, while maintaining lap times comparable to SAC and SAC-UPTrue's more aggressive policies (Table~\ref{table:multi_curriculum_perf}). Unlike \uptrue, which benefits from oracle access to the active mode but lacks a mechanism for sub-policy specialization, \model leverages its router to adaptively compose experts, enabling robust generalization. This carries over to the unseen Track 2 layout (Fig.~\ref{fig:racing_track_viz}\textbf{C}), where \model achieves the highest number of laps and lowest crash rate across \emph{all} surface assignments (Table~\ref{table:newoodtrack_perf}). In contrast, \uptrue, which was closest to \model on Track 1, falls significantly behind, in line with prior work~\cite{UPTrue} that mode-aware policies struggle in out-of-distribution settings. SAC matches \model in lap times but suffers up to 3x higher crash rates, indicating that aggressive performance alone is insufficient without robustness to mode switches.

\begin{table}[tb]
\centering
\footnotesize
\setlength{\tabcolsep}{5pt}
\setlength{\fboxsep}{1pt}
\resizebox{0.45\textwidth}{!}{
\begin{tabular}{lccc}
\toprule
\textbf{Eval. Surfaces} & \textbf{SAC} & \textbf{SAC-UPTrue} & \textbf{SAC-MoE (ours)} \\
\midrule
\multicolumn{4}{c}{\textbf{\textit{Single Surface}}} \\
\midrule
\textbf{\{\texttt{L}\}} & 0.3 / \textbf{30.2} / 0.98 & 0.1 / 34.4 / 1.00 & \textbf{0.5} / 33.3 / \textbf{0.94} \\
\textbf{\{\texttt{M}\}} & 2.2 / \textbf{30.5} / 0.21 & 0.3 / 32.1 / 0.99 & \textbf{2.3} / 32.7 / \textbf{0.07} \\
\textbf{\{\texttt{H}\}} & 2.3 / 31.0 / 0.15 & 0.3 / \textbf{30.4} / 0.97 & \textbf{2.5} / 31.9 / \textbf{0.00} \\
\midrule
\multicolumn{4}{c}{\textbf{\textit{Multi-Surface}}} \\
\midrule
\textbf{\texttt{\{L, M\}}} & 1.5 / \textbf{30.6} / 0.52 & 0.4 / 33.5 / 0.90 & \textbf{1.7} / 32.9 / \textbf{0.37} \\
\textbf{\texttt{\{L, H\}}} & 1.2 / \textbf{31.1} / 0.60 & 0.3 / 32.9 / 0.96 & \textbf{1.5} / 32.8 / \textbf{0.43} \\
\textbf{\texttt{\{M, H\}}} & \textbf{2.3} / \textbf{30.9} / 0.14 & 0.3 / 31.5 / 0.98 & \textbf{2.3} / 32.3 / \textbf{0.06} \\
\textbf{\texttt{\{L, M, H\}}} & 1.5 / \textbf{30.7} / 0.52 & 0.2 / 32.1 / 0.99 & \textbf{1.6} / 32.6 / \textbf{0.40} \\
\bottomrule
\end{tabular}
}
\caption{\small{\textbf{Quantitative racing results (Track 2).}
Entries show Laps Completed ($\uparrow$) / Avg. Lap Time (s) ($\downarrow$) / Crash Rate ($\downarrow$). \textbf{\texttt{L, M, H}} are low, med., high friction surfaces resp. For contexts across each set of surfaces, best results across models are \textbf{bolded}.}}
\label{table:newoodtrack_perf}
\vspace{-15pt}
\end{table}

\subsection{Case Study 2 - MuJoCo Simulations} \label{subsec:mujoco_results}
We evaluate policies in the Walker2d-v5 environment~\cite{towers2024gymnasium}, where agents are incentivized to walk quickly while maintaining the torso angle within safe bounds. As in Section~\ref{subsec:racetrack_results}, we introduce spatially varying friction. In addition, legged robots exhibit latent modes arising from internal joint configurations and gait patterns~\cite{gait_gp, DHA}. As a result, these modes depend on both spatial variation and internal state, and are only partially observable, even in simulation. Hence, the \uptrue, \sacsw baselines receive partial mode information (friction), but not full knowledge of the latent mode parameters, which are difficult to define without expert domain knowledge. This makes the task particularly challenging and motivates our study of whether \moe can apply to such systems with complex contact dynamics. 

\begin{figure}[t]
    \centering
    \includegraphics[width=0.8\linewidth]{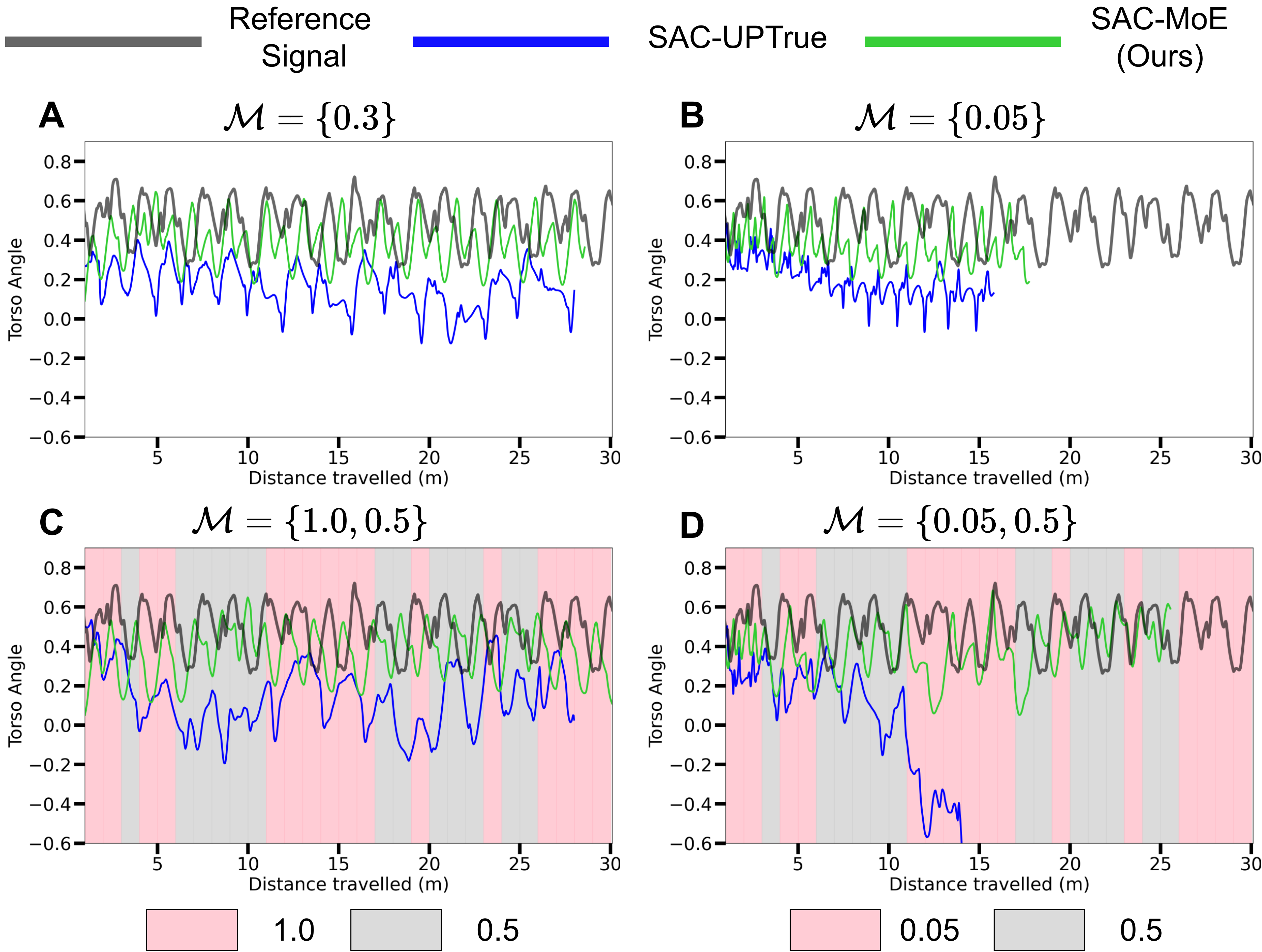}
    \caption{\small{\textbf{Torso-angle trajectories vs.\ friction context.} Comparison of \uptrue and \model across single- and multi- friction settings. Reference signal (black) is generated by an optimal SAC policy trained and tested \emph{only} in $\Mc = \{1.0\}$. Mode switching locations for \textbf{C}, \textbf{D} are visualized by colored boxes.}}
    \label{fig:torso_angle_viz}
    \vspace{-20pt}
\end{figure}

\noindent\textbf{Training and Evaluation setup.}  
Training is performed on with context space given by $\Ltrain = \{1.0, 0.1\}$ (friction assignments). All policies are trained for 1M steps using Curriculum~C from Section~\ref{subsec:racetrack_results}.
At test time, we evaluate generalization to unseen contexts $\Ctest \supset \Ctrain$ characterized by novel friction coefficients and mode switching locations (as visualized in Fig.~\ref{fig:torso_angle_viz}\textbf{C}). Episodes are over 1500 steps.
We report average distance travelled and torso velocity over 100 evaluation episodes per policy.  

\textbf{\textit{Result Discussion: }} Our results on the test \gls{hcmdp} are summarized in Table~\ref{table:walker_quant}. We structure the discussion in two parts, i) evaluating zero-shot generalization and qualitatively visualizing SAC-MoE's robustness to novel contexts (\textbf{P2}) ii) visualizing how the learned router activates different expert sub-policies corresponding to different latent modes to demonstrate how the \moe actor captures information about hybrid system dynamics implicitly without requiring context knowledge (\textbf{P3}). 

\noindent\textbf{Generalization performance (P2).}  
Table~\ref{table:walker_quant} highlights several trends. SAC, despite showing competitive zero-shot generalization in the racetrack task, now fails to maintain performance across contexts. \sacsw likewise struggles to compose its single-mode policies under switching. \uptrue performs best in high-friction settings but degrades sharply in harder single- and multi-friction contexts. In contrast, \model retains performance on increasingly difficult novel contexts, demonstrating robustness consistent with Section~\ref{subsec:racetrack_results}.

\begin{figure}[t]
    \centering
        \includegraphics[width=\linewidth]{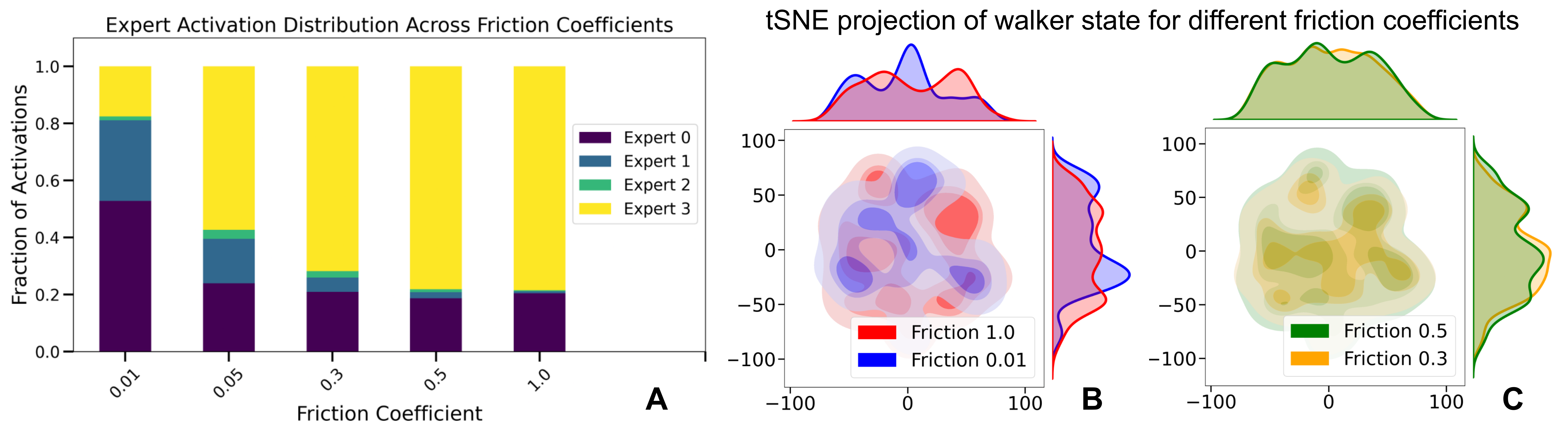}
    \caption{\small{\textbf{Correlation between expert activation and friction coefficient.} \textbf{(A)} We show the distribution of expert activations across friction parameters. 
    \textbf{(B,C)} We compute a t-SNE projection of the agent’s state vectors over an episode and plot their densities for different friction coefficients. 
    }}
    \label{fig:walker_qualit}
    \vspace{-15pt}
\end{figure}

Fig.~\ref{fig:walker_qualit} further illustrates this robustness by visualizing torso angle trajectories across contexts. Episodes terminate if the torso angle leaves $[-0.8, 0.8]$ radians, making this variable critical. As a reference, we train an SAC policy in a single high-friction environment ($\mu=1.0$) and evaluate it in the \emph{same} environment to get a near-optimal trajectory. Across all visualized contexts, \model tracks this reference more closely than \uptrue, thus resulting in up to 1.7x performance improvement in harder contexts.

\begin{table}[t!]
    \centering
    \setlength\tabcolsep{5pt}
    \scalebox{0.85}{
    \begin{tabular}{lcccc}
        \toprule
        \textbf{$\mathcal{M}$} 
        & \textbf{SAC} & \textbf{\uptrue} & \textbf{\sacswshort} & \textbf{\model (ours)} \\
        \midrule
        \multicolumn{5}{c}{\textbf{\textit{Single Friction}}} \\
        \midrule
        \textbf{\(\{1.0\}\)}  & 37.3 / 3.12 & 38.8 / 4.16 & \textbf{39.5 / 3.41} & 32.5 / 2.71 \\
        \textbf{\(\{0.3\}\)}  & 4.5 / 1.72  & \textbf{28.4 / 2.37} & 5.0 / 0.93                & 28.2 / 2.36 \\
        \textbf{\(\{0.1\}\)}  & 6.5 / 1.33  & 18.0 / 1.59 & 13.1 / 1.26          & \textbf{22.7 / 1.90} \\
        \textbf{\(\{0.05\}\)} & 6.9 / 0.96  & 13.8 / 1.20 & 7.5 / 0.84                    & \textbf{17.8 / 1.48} \\
        \midrule
        \multicolumn{5}{c}{\textbf{\textit{Multi-Friction}}} \\
        \midrule
        \textbf{\(\{1.0, 0.5\}\)}  & 8.8 / 2.26  & \textbf{38.1 / 3.72} & 6.2 / 2.52                & 31.8 / 2.70 \\
        \textbf{\(\{0.1, 1.0\}\)}  & 5.6 / 1.35  & 18.1 / 2.08 & 8.8 / 1.42           & \textbf{28.5 / 2.38} \\
        \textbf{\(\{0.1, 0.3\}\)}  & 4.8 / 1.24  & 20.3 / 1.78 & 14.6 / 1.34                    & \textbf{25.5 / 2.14} \\
        \textbf{\(\{0.05, 0.5\}\)} & 4.5 / 0.92  & 14.4 / 1.59 & 6.3 / 0.88                    & \textbf{24.2 / 2.04} \\
        \bottomrule
    \end{tabular}
    }
    \caption{\small \textbf{Quantitative results on Walker2D.} We report distance traveled (m) / torso velocity (m/s) for various models on single- and mixed-friction settings. Higher is better for both metrics; bold indicates the best \emph{distance} per row.}
    \label{table:walker_quant}
    \vspace{-10pt}
\end{table}

\noindent\textbf{Router expert activations and latent modes (P3).}
To address \textbf{P3}, we examine how router activations correlate with latent modes by tracking token assignments to experts under different friction coefficients.
Fig.~\ref{fig:walker_qualit}\textbf{A} shows that activations vary across friction contexts. A t-SNE projection of state observations shows that large friction differences (Fig.~\ref{fig:walker_qualit}\textbf{B}) yield distinct state distributions and correspondingly different expert activations, while similar friction (Fig.~\ref{fig:walker_qualit}\textbf{C}) produces overlapping distributions and similar activation patterns. In harder (lower-friction) contexts, activations spread across more experts, highlighting the benefit of the \moe router’s adaptive composition advantage over $\pisw$ (Section~\ref{subsec:moe_actor_rep}).

\section{Conclusion}
\glsreset{hcmdp}
We formalize the problem of learning policies for hybrid dynamical systems under uncertainty by introducing the \gls{hcmdp}. We highlight the challenges of controlling such systems with switching-based policies and address them by proposing \model capable of adaptively composing sub-policies. We develop a curriculum learning approach for systematic training across contexts, which consistently improves performance across all evaluated models. Through extensive simulations on autonomous racing and locomotion tasks, we show that \moe approach consistently outperforms baselines in zero-shot generalization to novel contexts.

\textbf{Limitations.} 
A key limitation of our zero-shot generalization setting is the requirement on the policy to adapt to unseen contexts during a single episode, despite observations early in the episode provide little information about potential future mode transitions. This makes our policy more conservative than necessary due to operation under uncertainty. 
Moreover while our adaptive curriculum improves training in finite context sets, scaling such approaches to larger context spaces remains a significant problem under the difficulties generated by \glspl{hcmdp} highlighted in Section~\ref{sec:curriculum_learning}.

\textbf{Future work.} 
Future work could involve considering a few-shot generalization setting to allow use of a latent context encoder by progressively refining context estimates across multiple episodes. This would serve to reduce policy conservatism by reducing context uncertainty while better capturing relations across the context space.
The above also allows better curriculum strategies that use latent encodings to capture similarities between contexts. This allows for treating contexts as more than just elements in $\Ctrain$ and addresses the problem highlighted in Section~\ref{sec:curriculum_learning}.

\bibliographystyle{ieee_nodate}

\bibliography{bibs/sacmoe}
\end{document}